\documentclass[conference]{IEEEtran}

\usepackage[utf8]{inputenc}
\usepackage[T1]{fontenc}
\usepackage{microtype}
\usepackage[ngerman,english]{babel}
\usepackage{listings} \usepackage{amsmath}
\usepackage[hidelinks]{hyperref}
\usepackage[capitalise]{cleveref}
\usepackage{multicol}
\usepackage{glossaries}
\glsdisablehyper
\usepackage[commandnameprefix=always,todonotes={textsize=tiny,obeyFinal}]{changes}
\usepackage{tikz}
\usetikzlibrary{positioning,shapes.multipart, math, shadows.blur, external}
\tikzset{
	panel/.style={inner sep=0, outer sep=0, anchor=north west},
	fig label/.style={label, outer sep=0, inner sep=1pt, anchor=north west, font=\small\bfseries},
}
\usepackage[binary-units=true]{siunitx}
\usepackage{dblfloatfix}
\usepackage{bm} \usepackage{mathtools}
\usepackage{physics}
\usepackage{csquotes}
\usepackage{graphicx}
\graphicspath{{./fig/}}
\usepackage{booktabs}
\usepackage{dcolumn} \newcolumntype{C}{D{,}{,\,}{2.2}}
\usepackage[backend=biber,maxbibnames=1,minbibnames=1,mincitenames=1,firstinits=true,sorting=none,doi=true,eprint=false,isbn=false,url=false]{biblatex}
\newacronym{abc}{ABC}{approximate Bayesian computation}
\newacronym{adc}{ADC}{analog-to-digital converter}
\newacronym{adex}{AdEx}{adaptive exponential integrate-and-fire}
\newacronym{bss2}{\mbox{BSS-2}}{Brain\mbox{ScaleS-2}}
\newacronym{cmos}{CMOS}{complementary metal-oxide semiconductor}
\newacronym{isi}{ISI}{interspike interval}
\newacronym{maf}{MAF}{masked autoregressive flow}
\newacronym{nde}{NDE}{neural density estimator}
\newacronym{resnet}{ResNet}{residual network}
\newacronym{sbi}{SBI}{simulation-based inference}
\newacronym{snpe}{SNPE}{sequential neural posterior estimation}
\newacronym{vae}{VAE}{variational autoencoder}
 
\newlength{\singlecolumn}
\newlength{\doublecolumn}
\newcommand{\myvec}[1]{\bm{#1}}

\newcommand{\captiontitle}[1]{\textbf{#1. }}
\newcommand{\formatpanel}[1]{(\lowercase{#1})}
\newcommand{\subcaption}[1]{\textbf{\formatpanel{#1}}}

\makeatletter
\newcommand{\@giventhatstar}[2]{\left(#1\;\middle|\;#2\right)}
\newcommand{\@giventhatnostar}[3][]{#1(#2\;#1|\;#3#1)}
\newcommand{\giventhat}{\@ifstar\@giventhatstar\@giventhatnostar}
\makeatother

\renewbibmacro{finentry}{\iffieldequalstr{entrykey}{radev2023bayesflow}{\finentry\newpage}{\finentry}}

\newcommand\copyrighttext{\footnotesize This article has been accepted for presentation at the \textit{Neuro Inspired Computational Elements Conference 2026} and will appear in the conference proceedings.\\
	\textcopyright 2026 IEEE. Personal use of this material is permitted.
  Permission from IEEE must be obtained for all other uses, in any current or future
  media, including reprinting/republishing this material for advertising or promotional
  purposes, creating new collective works, for resale or redistribution to servers or
  lists, or reuse of any copyrighted component of this work in other works.}
\newcommand\copyrightnotice{\begin{tikzpicture}[remember picture,overlay]
\node[anchor=south,yshift=10pt] at (current page.south) {\fbox{\parbox{\dimexpr\textwidth-\fboxsep-\fboxrule\relax}{\copyrighttext}}};
\end{tikzpicture}}

\begin{document}

\title{Amortized Inference of Neuron Parameters on Analog Neuromorphic Hardware}

\author{\IEEEauthorblockN{Jakob Kaiser\IEEEauthorrefmark{1}\IEEEauthorrefmark{2},
	Eric Müller\IEEEauthorrefmark{3}
	and
	Johannes Schemmel\IEEEauthorrefmark{2}}\IEEEauthorblockA{\IEEEauthorrefmark{2}\textit{Institute of Computer Engineering}, Heidelberg University, Germany}\IEEEauthorblockA{\IEEEauthorrefmark{3}\textit{Kirchhoff Institute for Physics},   Heidelberg University, Germany}\IEEEauthorblockA{\IEEEauthorrefmark{1}\href{mailto:jakob.kaiser@kip.uni-heidelberg.de}{jakob.kaiser@kip.uni-heidelberg.de}}}

\maketitle
\copyrightnotice

\begin{abstract}
Our work utilized a non-sequential \acrlong{sbi} algorithm to provide an amortized \acrlong{nde}, which approximates the posterior distribution for seven parameters of the \acrlong{adex} neuron model of the analog neuromorphic \acrlong{bss2} substrate.
We constrained the large parameter space by training a binary classifier to predict parameter combinations yielding observations in regimes of interest, i.e.\ moderate spike counts.
We compared two \acrlongpl{nde}:
one using handcrafted summary statistics and one using a summary network trained in combination with the \acrlong{nde}.
The summary network yielded a more focused posterior and generated posterior predictive traces that accurately captured the membrane potential dynamics.
When using handcrafted summary statistics, posterior predictive traces match the included features but show deviations in the exact dynamics.
The posteriors showed signs of bias and miscalibration but were still able to yield posterior predictive samples that were close to the target observations on which the posteriors were constrained.
Our results validate amortized \acrlong{sbi} as a tool for parameterizing analog neuron circuits.
 \end{abstract}

\begin{IEEEkeywords}
Neuromorphic Computing, Simulation-Based Inference, AdEx, BayesFlow
\end{IEEEkeywords}

\section{Introduction}\label{sec:introduction}
\glsresetall

Identifying suitable models to replicate physical observations is often a laborious process.
Even when a model is available, determining an effective parameterization can be challenging.
Various methods have been used in neuroscience to achieve appropriate model parameterizations.
These methods include directly deriving model parameters by fitting experimental observations, genetic algorithms, gradient-based methods, and \gls{sbi} \cite{hertaeg2012approximation,vanier1999comparative,vangeit2008automated,deistler2024differentiable,goncalves2020training}.

In recent years, neural \gls{sbi} methods have established themselves as powerful tools for approximating parameter distributions for complex models and have been used for neuroscientific problems as well \cite{goncalves2020training,deistler2022truncated,schmitt2024consistency,gloeckler2024all}.
In these methods, a \gls{nde}, a flexible neural network, is trained to approximate the posterior distribution of model parameters $\myvec{\theta}$ given a target observation $\myvec{x}$.

In previous studies, we employed a sequential \gls{sbi} algorithm, the \gls{snpe} algorithm, to parameterize multi-compartmental neuron models as well as \gls{adex} neurons on \gls{bss2} \cite{kaiser2023simulation,huhle2024reproduction}.
We showed that the \gls{snpe} algorithm can deal with the temporal noise present in the mixed-signal \gls{bss2} system.
A sequential algorithm promises more informative samples in later inference rounds \cite{greenberg2019automatic} but comes at the cost of a specialized \gls{nde}: it can only approximate a posterior for a single target observation that was used during training and does not amortize over arbitrary target observations.

In the current study, we consider a non-sequential \gls{sbi} algorithm that provides an amortized \gls{nde}.
This allowed for the inference of model parameters for many different observations and a more thorough evaluation of the posterior approximation.
Furthermore, we explored two different approaches to extract relevant features from the experimental recordings.
In one case, we relied on handcrafted summary statistics, which were calculated from the spike times and membrane recordings.
In the other case, we trained a neural network to extract relevant features from the membrane recordings alone.

\subsection{The \acrlong{bss2} System}\label{sec:intro:bss2}

The neuromorphic \gls{bss2} system combines analog circuits for the emulation of synapse and neuron dynamics with digital logic for handling event communication and configuration \cite{pehle2022brainscales2,billaudelle2022accurate}.
\Gls{bss2} implements the dynamics of the \gls{adex} neuron model \cite{brette2005adaptive} in \num{512} neuron circuits \cite{billaudelle2022accurate}.
The dynamics of the \gls{adex} model are described by two coupled differential equations for the membrane potential $V_\text{m}$ and adaptation current $w$.
On \gls{bss2}, both are emulated in continuous time using analog circuits:
\begin{equation}
	\label{eq:adex}
	\begin{split}
		C_\text{m} \dv{V_\text{m}(t)}{t} & = g_\text{l} \cdot \left( V_\text{l} - V_\text{m}(t) \right) \\
		                                 & + g_\text{l} \Delta_\text{T} \exp\left( \frac{V_\text{m}(t) - V_\text{T}}{\Delta_\text{T}} \right) \\
										 & + I_\text{syn}(t) + I(t) - w(t),
	\end{split}
\end{equation}
with the membrane capacitance $C_\text{m}$, leak conductance $g_\text{l}$, leak potential $V_\text{l}$, exponential slope factor $\Delta_\text{T}$ and effective threshold $V_\text{T}$.

Here, $I_\text{syn}$ represents the synaptic input current and $I(t)$ an arbitrary external current input.
The evaluation of the adaptation current $w$ is given by
\begin{equation}
	\label{eq:adaptation}
		\tau_\text{w} \dv{w(t)}{t} = a \left( V_\text{m}(t) - V_\text{l} \right) - w(t),
\end{equation}
where $a$ is the sub-threshold adaptation and $\tau_\text{w} = C_\text{w} / g_{\tau_\text{w}}$ the adaptation time constant, which is given by the capacitance $C_\text{w}$ and the conductance $g_{\tau_\text{w}}$.

The differential equations are complemented by two jump conditions.
Once the membrane potential $V_\text{m}$ reaches a hard threshold $V_\text{th}$, the membrane is reset to the reset potential $V_\text{r}$ for the duration of the refractory period $\tau_\text{ref}$.
Additionally, the adaptation current $w$ is increased by the spike-triggered adaptation increment $b$: $w \rightarrow w + b$.

The \gls{bss2} system can be run at an adjustable, accelerated speed; in this study, we selected an acceleration factor of \num{1000} relative to biological time.
Consequently, the characteristic timescales are on the order of \si{\us} instead of \si{\ms}, and the firing rates fall within the \si{\kHz} range.

A capacitor-based memory array converts digital values in the range of \numrange{0}{1022} to bias currents and voltages \cite{hock13analogmemory}.
This enables the individual configuration of all conductances and voltages introduced in \cref{eq:adex,eq:adaptation}.
The membrane capacitance $C_\text{m}$ and refractory period $\tau_\text{ref}$ can be controlled digitally.
\textcite{billaudelle2022accurate} discusses the implementation of the analog neuron circuits in more detail.

\subsection{Simulation-based Inference}

\Gls{sbi} aims to construct an approximation of the posterior distribution based on simulation results.
Given a model $\mathcal{M}: \Theta \rightarrow \mathcal{X}$ which maps parameters $\myvec{\theta} \in \Theta$ to observations $\myvec{x} \in \mathcal{X}$, \gls{sbi} tries to approximate the posterior distribution $p\giventhat{\myvec{\theta}}{\myvec{x}}$.

Several algorithms have been designed to approximate the posterior distribution \cite{sisson2018handbook,greenberg2019automatic,cranmer2020frontier,radev2020bayesflow}.
Recently, neural \gls{sbi} algorithms have been developed that utilize deep neural networks, also known as \glspl{nde}, to estimate the posterior distribution.
In this process, parameters $\{\myvec{\theta}_i\}$ are sampled from a prior distribution $\{\myvec{\theta}_i\} \sim p(\myvec{\theta})$, and observations $\{\myvec{x}_i\}$ are produced using the model $\mathcal{M}$ to create a training dataset $\{(\myvec{\theta}, \myvec{x})_i\}$.
This dataset is then used to train the \gls{nde} to approximate the posterior distribution.

In this study, we utilized an algorithm called \enquote{BayesFlow}~\cite{radev2020bayesflow}.
This algorithm enables the integration of a summary network $\mathcal{S}: X \rightarrow Y$ with the \gls{nde} to automatically derive features from potentially high-dimensional simulation data.
Here, $Y$ represents the space of summary statistics, which is generally of lower dimensionality than the observation space $X$.
The \gls{nde} is trained using a loss function based on the Kullback-Leibler divergence between the true and approximated posterior.
 
\section{Methods}\label{sec:methods}

We first describe the experimental setup and electrophysiological features extracted from the experimental recordings.
Next, we describe how we generated a dataset that can be used to train the \gls{nde}.
In the last section, we elaborate on how the \gls{nde} was trained.

\subsection{Experiment Design and Data Processing}\label{sec:methods:experiment}

We stimulated a single \gls{adex} neuron on the \gls{bss2} system with a step current.
The step current started at \SI{0.3}{\ms} and lasted for \SI{1}{\ms}.
We recorded the membrane potential $V_\text{m}(t)$ and spikes $S(t)$ for a total of \SI{1.6}{\ms} to include the membrane response after the step current was turned off.

Based on the work of \textcite{druckmann07,gouwens2018systematic}, we extracted several electrophysiological features from the recorded spike times $S(t)$ and membrane potential $V_\text{m}(t)$:
(1) the average firing rate $r$,
(2) the latency to the first spike $\Delta t_\text{spike}^\text{first}$ relative to the stimulus onset,
(3) the first \gls{isi} $\text{ISI}_\text{first}$,
(4) the last \gls{isi} $\text{ISI}_\text{last}$,
(5) the coefficient of variation of the \glspl{isi} $\text{CV}_\text{ISI}$,
(6) the average \gls{isi} $\bar{\text{ISI}}$,
(7) the adaptation index $A$,
(8) the baseline voltage $V_\text{0}$ at the experiment beginning,
(9) the minimum voltage $V_\text{min}$ after stimulus offset, 
(10) the fast trough depth $V_\text{FT}$, 
(11) the slow trough depth $V_\text{ST}$, 
and (12) the time of the slow trough $\Delta t_\text{ST}$ as a fraction of the first \gls{isi}.

Furthermore, to reduce the amount of data saved to disk, we interpolated the recorded membrane potentials $V(t)$ on a regular time grid with \num{10000} data points.

\subsection{Dataset}

We chose to alter a total of \num{7} parameters for which we wanted to determine the posterior distribution.
The parameters we consider are the leak conductance $g_\text{l}$, reset potential $V_\text{r}$, exponential slope factor $\Delta_\text{T}$, effective threshold $V_\text{T}$, subthreshold adaptation $a$, spike-triggered adaptation $b$, and conductance controlling the time constant of the adaptation $g_{\tau_\text{w}}$.

The other parameters were fixed.
\Cref{eq:adex} is invariant to a shift in voltages; therefore, we can keep one of the voltages fixed.
We chose to keep the leak potential $V_\text{l}$ fixed.
The hard threshold $V_\text{th}$ has only a minor influence on the neuron behavior \cite{clopath2007predicting}; therefore, we kept it fixed as well.
As we kept the maximum input amplitude of the current $I_\text{max}$ fixed, the membrane capacitance mainly scaled the membrane time constant $\tau_\text{m} = C_\text{m} / g_\text{l}$.
Because we already included the leak conductance $g_\text{l}$ in the parameters we altered, we kept the membrane capacitance $C_\text{m}$ constant.

To generate the dataset, we drew random parameterizations from a uniform distribution that covered the entire configurable range of the hardware, that is, each parameter could be in the range from \numrange{0}{1022}.
For an initial dataset, we drew \num{50000} pairs of parameters ${\myvec{\theta}_i}$, recorded the membrane potentials $V_\text{m}(t)$ as well as spike times $S(t)$ and computed the summary statistics.
We then saved the interpolated membrane trace, exact spike times, and observations in a dataset.

\subsubsection{Constraining the Parameter Space}\label{sec:methods:constrained}

Due to the large parameter space, a large set of parameters yields observations that do not show spiking behavior or a high number of spikes.
We want to exclude some of these parameter combinations and therefore train a binary classifier, as in \textcite{deistler2022energy}, on the initial dataset, which predicts whether a parameter combination $\myvec{\theta}$ yields a spike count in the range from \numrange{1}{70}.

Following \textcite{deistler2022energy}, we chose a \gls{resnet} classifier \cite{he2016deep}.
Our network consists of four blocks with \num{100} hidden features each; we used ReLU activation functions and a dropout rate of \num{0.5} during training.

After training the classifier, we used it to generate a dataset, which was then used to train the \gls{nde}.
As before, we started by drawing parameters from a uniform distribution.
However, this time, we only included parameters $\myvec{\theta}$ for which the classifier predicted a high probability that the observation is in the specified range.
We performed experiments for these parameters and collected the data in our dataset.

We used this approach to generate one dataset with \num{600000} samples for training the \gls{nde} and another dataset with \num{600} samples for validation.

\subsection{Training a Neural Density Estimator}

We utilized the Python library \texttt{Bayesflow} version \texttt{2.0.7} to train our \glspl{nde} \cite{radev2020bayesflow,radev2023bayesflow}.

The first \gls{nde} was trained using the handcrafted summary statistics introduced in \cref{sec:methods:experiment}.
This \gls{nde} is a coupling flow comprising \num{10} coupling blocks \cite{radev2020bayesflow}, and we trained the network for \num{150} epochs.

We trained a second \gls{nde} in conjunction with a summary network.
This \gls{nde} was also a coupling flow, but with \num{8} blocks. The summary network, based on \textcite{zhang2023solving}, consists of two convolutional neural networks followed by a recurrent neural network.
The kernel size, stride, and number of filters were (8, 4, 16) and (4, 2, 8) for the first and second convolutional layers, respectively.
The recurrent neural network comprised \num{128} units, which were connected via a dense layer to an output layer of size \num{14}.
We trained the summary network and \gls{nde} together for \num{30} epochs.
 
\section{Results}\label{sec:results}

First, we show the effect of the binary classifier on the spike distribution in the generated dataset and show sample traces from this dataset.

We then trained two \glspl{nde}: one with handcrafted summary statistics and one that included a summary network that was trained in combination with the \gls{nde}.
To ease the interpretation of the results, we first consider a single target observation $\myvec{\theta_\text{T}}$.
For this observation, we show the trial-to-trial variation due to temporal noise before examining the approximated posteriors and posterior predictive traces.
Furthermore, we performed a posterior predictive check based on the handcrafted summary statistics introduced in \cref{sec:methods:experiment}.

Finally, we show the posterior predictive traces for random observations of the validation dataset.

\subsection{Datasets}
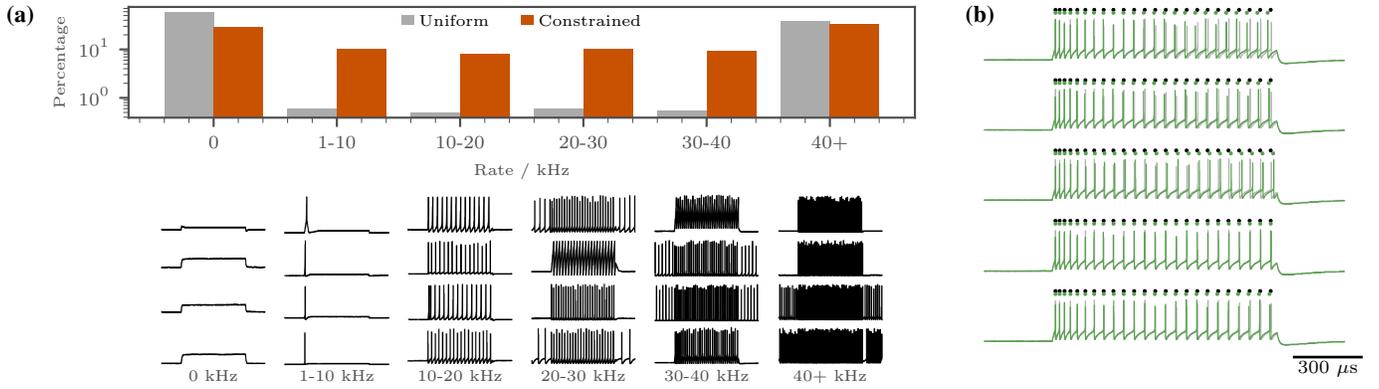
\begin{figure*}
	\newlength{\width}
	\setlength{\width}{0.7\doublecolumn}

	\begin{tikzpicture}
		\coordinate (a) at (0,0);
		\node[panel] at (a) {\includegraphics{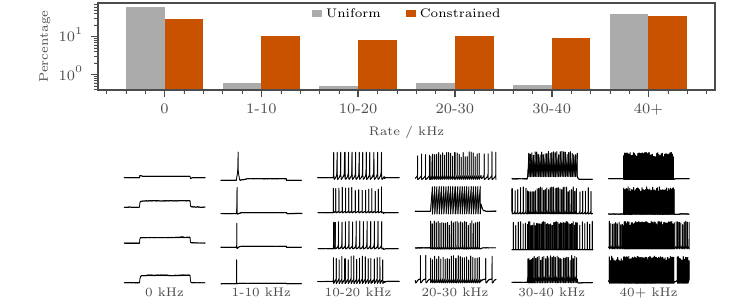}};
		\node[fig label] at (a) {\formatpanel{A}};
		\coordinate (b) at (\width, 0);
		\node[panel] at (b) {\includegraphics{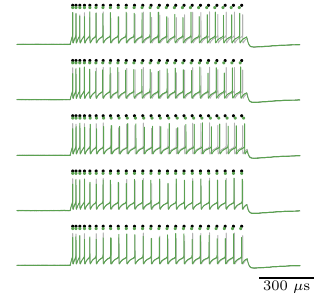}};
		\node[fig label] at (b) {\formatpanel{B}};
	\end{tikzpicture}
	\caption{\captiontitle{Target Observation and Dataset}
		\subcaption{A} The top panel shows the distribution of firing rates for the uniform dataset as well as for the dataset which was constrained with a binary classifier.
		For the uniform dataset, only a small fraction of samples had a firing rate in the range between \SIrange{1}{40}{\kilo\hertz}.
		With the help of a binary classifier, the number of samples with a firing rate in this range can be increased.
		The sample traces in the bottom panel show the diversity of observed firing patterns.
		\subcaption{B} Trial-to-trial variation at the target parameterization $\myvec{\theta_\text{T}}$.
		We display the observation $\myvec{x_\text{T}}$ which is used in \cref{sec:res:posterior} for comparison in gray.
		The spike times are indicated by dots.
		The experiment was repeated several times with the same parameterization $\myvec{\theta_\text{T}}$.
		The traces closely match each other at the beginning of the stimulus.
		As the experiments progressed, temporal noise resulted in noticeable differences in the recordings.
	}
  \label{fig:dataset}
\end{figure*}

The large configuration space of the experiment led to a large number of different responses in the dataset.
\Cref{fig:dataset}~\formatpanel{A} shows how the firing rates are distributed in the dataset.
When drawing the parameters from a uniform distribution, approximately \SI{98}{\percent} of the experiments showed either no spikes or more than \num{40} spikes during the experimental period.

Because we were mostly interested in responses with moderate spike counts, we trained a binary classifier, as described in \cref{sec:methods:constrained}, to increase the number of samples in the region of interest.
This increased the number of samples with a firing rate in the range between \SIrange{1}{40}{\kilo\hertz} from \SI{2}{\percent} to approximately \SI{38}{\percent}.

The exemplary traces in \cref{fig:dataset}~\formatpanel{A} show that the parameterization does not only affect the firing rate but also has an effect on spike widths and after-spike potentials.
Similar to our target trace, \cref{fig:dataset}~\formatpanel{B}, some of the traces show accommodation, for example, the third trace in the \SIrange{10}{20}{\kilo\hertz} range.

In some cases, the neuron spiked before the stimulus was applied.
In these cases, an increased firing rate during stimulus presentation can still be observed.
The last two traces in the last displayed category shortly stopped firing after the stimulus offset, indicating adaptive behavior.

\subsection{Trial-to-Trial Variation}

\Cref{fig:dataset}~\formatpanel{B} shows five experiment repetitions at the target parameterization $\myvec{\theta_\text{T}}$.
We chose this parameterization such that we see a moderate number of spikes and an accommodating behavior, that is, the \gls{isi} increases over time.
The voltage levels and general evolution of the membrane potential were similar in all recordings; however, spike timings varied from trial to trial.
While the spikes align well at the beginning of the stimulus, the mismatch between the individual spike times increases as the experiment progresses.

\subsection{Approximated Posterior of a Example Observation}\label{sec:res:posterior}
\begin{figure*}
	\newlength{\halfwidth}
	\setlength{\halfwidth}{0.6\doublecolumn}

	\begin{tikzpicture}
		\coordinate (a) at (0,0);
		\node[panel] at (a) {\includegraphics{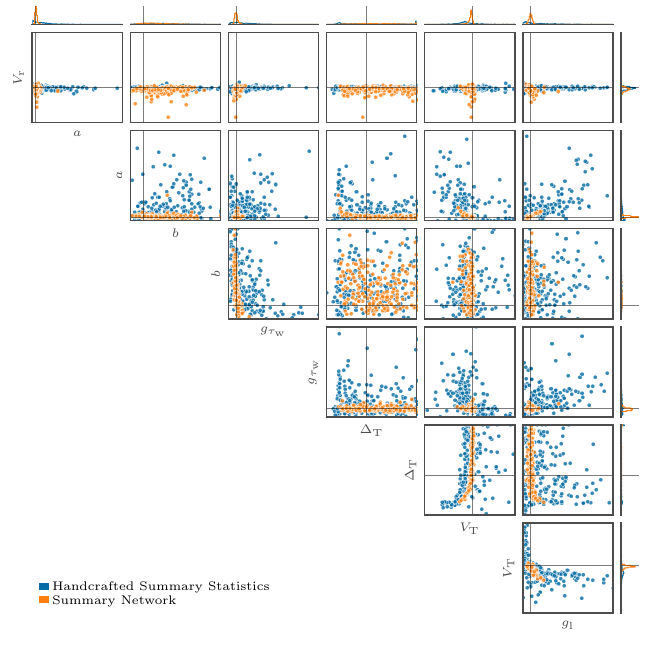}};
		\node[fig label] at (a) {\formatpanel{A}};
		\coordinate (b) at (\halfwidth, 0);
		\node[panel] at (b) {\includegraphics{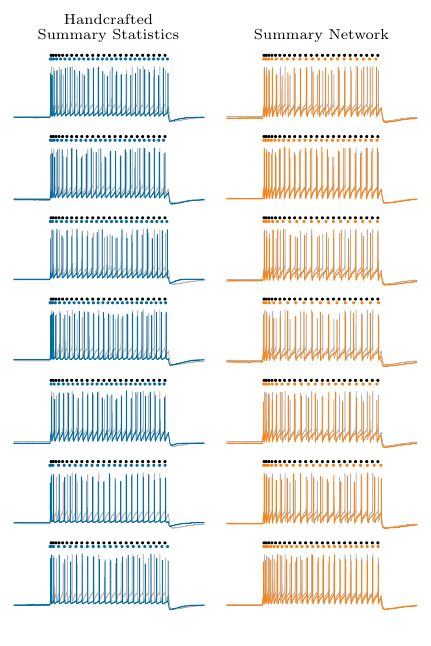}};
		\node[fig label] at (b) {\formatpanel{B}};
	\end{tikzpicture}
	\caption{\captiontitle{Approximated Posterior and Posterior Predictive Samples.}
		\subcaption{A} Two- and one-dimensional distributions of samples drawn from the approximated posteriors.
		The parameter set $\myvec{\theta_\text{T}}$ used to generate the target observations $\myvec{x_\text{T}}$ is indicated by lines.
		The axes span the entire range of the uniform prior distribution from \numrange{0}{1022}.
		The posterior trained with the handcrafted summary statistics (blue) and the one trained with a summary network (orange) both show higher sample densities near the target parameters $\myvec{\theta_\text{T}}$.
		When using the summary network, the samples are more focused on these target parameters.
		A strong correlation between the exponential slope factor $\Delta_\text{T}$ and the exponential threshold $V_\text{T}$ can be observed.
		The values of the spike-triggered adaptation $b$ are for both approximations spread over a broad range.
		\subcaption{B} For both posterior approximations, we drew \num{1000} random samples and plot the membrane response for the parameters with the highest posterior probability.
		While the exact spike timings vary, the overall spiking behavior of the posterior predictive samples is similar to the target observation: all samples show accommodation and have a similar spike count.
		In the case of the handcrafted summary statistics, the exact evolution of the membrane potential between spikes varies considerably from the target trace.
		When using a summary network, the voltage evolution more closely matches the target trace.
	}
  \label{fig:posterior}
\end{figure*}
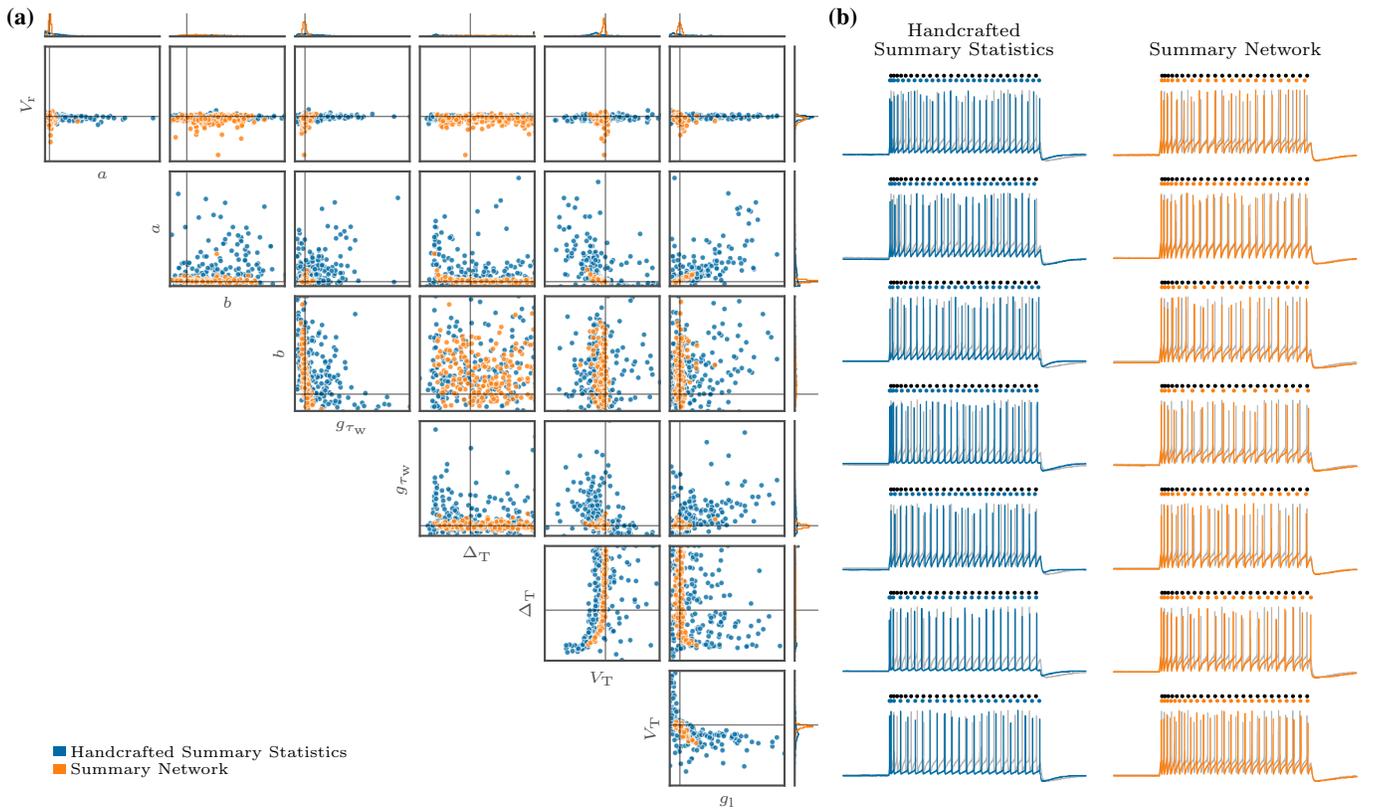

In this section we illustrate the approximated posterior for a single target observation $\myvec{x_\text{T}}$ which was generated with known parameters $\myvec{\theta_\text{T}}$, \cref{fig:posterior}~\formatpanel{A}.
When using the handcrafted summary statistics (see \cref{sec:methods:experiment}), the posterior samples still cover a considerable range of the parameter space but show increased densities near the target parameters $\myvec{\theta_\text{T}}$.
For the reset potential $V_\text{r}$ the samples are the most focused, while the samples are spread over a broad range for the spike-triggered adaptation $b$.

In the case of the summary network, the samples are more focused on the target parameters $\myvec{\theta_\text{T}}$.
Similar to the handcrafted summary statistics, a strong correlation between the exponential slope factor $\Delta_\text{T}$ and the exponential threshold $V_\text{T}$ can be observed.
Furthermore, the spike-triggered adaptation $b$ once more spreads over a broad range.

Next, we drew \num{1000} random samples from the individual posteriors, selected those with the highest posterior probability, and ran the experiment with these parameters; this yielded posterior predictive traces as shown in \cref{fig:posterior}~\formatpanel{B}.
For the handcrafted summary statistics, the posterior predictive traces showed a similar spiking behavior as the target trace: all the traces showed accommodation, and the spike counts were comparable.
However, the first spike occurred earlier in the posterior predictive samples than in the target trace.
In addition, the evolution of the membrane potential showed clear deviations between the target observation and posterior traces.
This difference is most visible between spikes.
Here, most of the posterior traces showed a flatter evolution with a rapid increase in voltage just before spiking.

When using the summary network, the spiking behavior still matched the target behavior.
However, the evolution of the membrane potential followed the target trace more closely.

\begin{figure*}

	\begin{tikzpicture}
		\coordinate (a) at (0,0);
		\node[panel] at (a) {\includegraphics{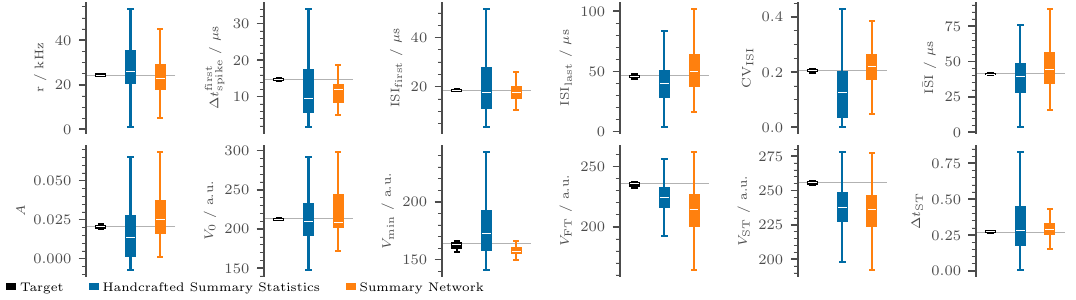}};
	\end{tikzpicture}
	\caption{\captiontitle{Posterior Predictive Check}
	We used handcrafted summary statistics to compare the predictive posterior samples with the target observation.
	We drew \num{1000} parameters $\myvec{\theta}$ from each posterior distribution and perform an experiment to record the observations $\myvec{x}$.
	In the case of the target, we used the same parameterization $\myvec{\theta_\text{T}}$ for all experiments.
	The boxes span the range from the first to the third quartile, and the whiskers span the farthest data point within the 1.5 interquartile range.
	The horizontal white lines mark the median value; the horizontal gray line indicates the features of the target observation $\myvec{x}_\text{T}$.
	For most features, the posterior predictive samples show a similar behavior as the target.
	Some features such as the time to first spike $\Delta t_\text{spike}^\text{first}$ or the fast trough depth $V_\text{FT}$ as well slow trough depth $V_\text{ST}$ are underestimated in the posterior predictive samples.
	The variation in the posterior predictive samples is considerably larger than the trial to trial variations.
	}
  \label{fig:ppc}
\end{figure*}

Next, we performed a posterior predictive check.
For this, we calculated the handcrafted summary statistics, see \cref{sec:methods:experiment}, for trials at the target parameterization $\myvec{\theta_\text{T}}$ and for posterior predictive samples, \cref{fig:ppc}.
Most of the target features lie within the first to third quartiles of the corresponding posterior predictive samples of the posterior trained using the handcrafted summary statistics.
The fast trough depth $V_\text{FT}$ and slow trough depth $V_\text{ST}$ target features do not fall within this range and are lower than the target.
As can already be seen in the posterior predictive traces in \cref{fig:posterior}~\formatpanel{B}, the firing rate $r$ and the mean \gls{isi} $\bar{\text{ISI}}$ are recovered successfully.
This is also true for the other quantities related to the \gls{isi} and the time of the slow trough $\Delta t_\text{ST}$.
Compared to the trial-to-trial variation, the variation in the posterior predictive samples is considerably higher.

This variation is reduced for about half of the features when a summary network is used.
Since the posterior predictive observations are not always centered around the value of the target observation, for more features the target observation does not fall within the first to third quartiles.
Similar to the posterior predictive samples of the posterior trained with the handcrafted features, the quantities related to the rate $r$ and \gls{isi} could be recovered well, whereas the fast trough depth $V_\text{FT}$ and slow trough depth $V_\text{ST}$ were underestimated.
Furthermore, the time to the first spike $\Delta t_\text{spike}^\text{first}$ is underestimated as well.

\subsection{Amortized Posterior}

\begin{figure*}

	\begin{tikzpicture}
		\coordinate (a) at (0,0);
		\node[panel] at (a) {\includegraphics{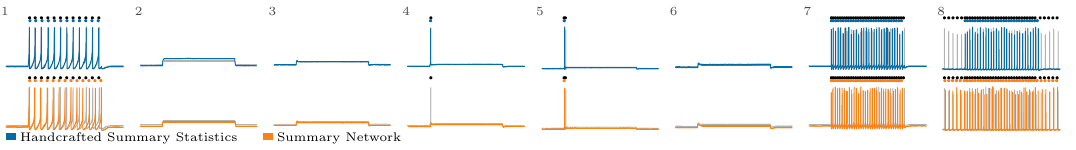}};
	\end{tikzpicture}
	\caption{\captiontitle{Posterior Predictive Traces}
	We drew \num{8} random observations from the validation set and plotted the posterior predictive trace with the highest probability out of \num{10000} draws from the corresponding posterior distribution.
	The gray traces indicate the target observations, and the spikes are displayed as dots.
	The posterior based on handcrafted summary statistics (top/blue) can recover most features of the target traces.
	If no spikes are present or if the target observation includes spikes outside the stimulus region, the posterior predictive traces vary significantly from the target.
	When a summary network (bottom/orange) is used, these posterior predictive traces closely resemble the target traces.
	For the fourth observation, the posterior predictive trace of the posterior which includes a summary network does not spike, whereas the target observation spikes once.
	}
  \label{fig:amortized}
\end{figure*}

In the previous section, we evaluated the posterior approximation for a single observation $\myvec{x_\text{T}}$.
Because we trained an amortized \gls{nde}, we can generate posterior approximations for arbitrary observations.
In \cref{fig:amortized}, we draw random observations from a validation dataset, draw \num{10000} random parameterizations from the corresponding posterior distributions, and use the most probable one to generate a posterior predictive sample.

Overall, both \glspl{nde} can produce traces that closely match most given target traces.
In the case of handcrafted summary statistics, the posterior predictive traces show variations if there are no spikes or if the neuron spikes outside the stimulus region.
The \gls{nde} which includes the summary network performs better in these cases but fails in another case in which the target observation spikes once.

Similarly, both posterior predictive traces shows one spike for the fifth trace, whereas the target observation includes two spikes in quick succession.
 
\section{Discussion}\label{sec:discussion}
\glsresetall

We trained \glspl{nde} to approximate the posterior distribution of seven parameters of a \gls{adex} neuron emulated on \gls{bss2}.
To train the \glspl{nde}, we first generated a dataset with parameters drawn from a uniform distribution over the entire parameter space and filtered using a binary classifier, which was trained on a smaller dataset beforehand.

We trained two different \glspl{nde}: one relying on handcrafted summary statistics and the other including a summary network that was trained in conjunction with the \gls{nde}.
Next, we illustrated the approximated posterior distribution, high-probability posterior predictive traces, and posterior predictive check for a single example observation.
We concluded by showing high-probability posterior predictive traces for other unseen observations from the validation dataset.

\subsection{Constrained Dataset}
Due to the vast configuration space of the \gls{bss2} system, many possible parameterizations result in behaviors that are not of interest to us.
Inspired by \textcite{deistler2022energy}, we trained a binary classifier to detect parameterizations that lead to interesting observations (here defined as a firing rate in the range from \SIrange{1}{70}{\kilo\hertz}).
Based on this classifier, we selected the parameterizations used to generate a dataset that yielded a much higher number of samples in the region of interest.

While we want to decrease the number of these uninteresting samples, we still want to include some of them in our training dataset to explore a larger parameter space.
\textcite{radev2020bayesflow} argue that these, less informative samples for a target observation, can still benefit the training of the \gls{nde}.
We explicitly excluded non-spiking observations in the acceptable range of the classifier, but chose a relatively high rejection threshold to still include some of these samples and have a false negative rate, that is, the rate of rejecting parameterizations that yield interesting observations, below \SI{5}{\percent}.

With this we were able to increase the percentage of samples in the range from \SIrange{1}{40}{\kilo\hertz} from about \SI{2}{\percent} to \SI{38}{\percent}.

\subsection{Posterior Approximations}

We visualized the posterior approximations given an exemplary observation for both \glspl{nde}, one using handcrafted summary statistics and the other using a summary network.
For this particular observation, both \glspl{nde} yielded posterior densities that were high in the vicinity of the parameter settings used to generate the target observation.
The \gls{nde} which included the summary network showed a much more focused posterior distribution.
This is expected because it potentially captures more features of the generated membrane potential than the handcrafted summary features \cite{kaiser2023simulation}.

This is also evident in the high-probability posterior predictive traces.
Here, the \gls{nde} trained on the handcrafted summary statistics was able to produce samples that closely matched the features that were part of the summary statistics, such as the baseline voltage, \gls{isi}, or spike count.
However, the posterior predictive traces failed to capture the exact dynamics of the membrane potential, and the evolution of the membrane potential between the spikes appeared significantly different.
Because the \gls{nde} with the summary network had access to the entire membrane recording, its posterior predictive traces followed the target trace more closely.

We then used handcrafted summary statistics to conduct a posterior predictive check.
In general, the posterior predictive samples of both \gls{nde} were near the exemplary target observation.
However, for some features clear deviations were visible.
In particular, the fast and slow trough depths and the latency to the first spike were underestimated in the posterior predictives.

Finally, we show the posterior predictive traces for other random observations from the validation dataset.
In most cases, the \glspl{nde} reproduced observations similar to the corresponding target observations.
The handcrafted summary statistics heavily rely on spike features during the stimulus, such that the \gls{nde} trained with these statistics struggled when no spikes were present or when the neuron spikes even when no stimulus is presented.
In these cases, the \gls{nde} which includes the summary network performed better.

However, it failed in another case in which the target trace spiked once, but the posterior predictive trace did not show any spikes.
We emulated the target parameterization several times for this trace and observed that in a few cases, the target configuration also produced non-spiking results (4 out of 100 tries).
This indicates that the configuration is in a regime in which the response is more susceptible to temporal noise.
As the summary network considers the entire membrane recording, it does not necessarily optimize the spike count but rather the overall behavior.

Similarly, in another case where the target trace deviated from the predictive samples, we observed that the same parameterization yielded in \num{24} out of \num{100} trials observations with one spike and in all other cases two spikes..
This suggests that the configuration is once again more susceptible to temporal noise, making it more difficult to produce an exact match.

\subsection{Conclusion and Outlook}

The \gls{adex} neuron can produce many different behaviors when a large parameter space is considered.
Nevertheless, the \glspl{nde} trained in this study were able to produce posterior predictive traces that closely resembled the target traces.
The mismatch between the target and the predictive traces is on one hand due to the intrinsic temporal noise on analog neuromorphic hardware.
On the other hand, our results suggest that the posterior approximations might be biased and not well calibrated.
In future work, we will conduct a more comprehensive assessment of the calibration properties of the inferred posterior distributions and develop methods to enhance their accuracy.
In recent years, diffusion-based models have been increasingly adopted within the framework of \gls{sbi} \cite{gloeckler2024all,schmitt2024consistency}; we intend to investigate the suitability and performance of these models for our specific inference problem.
Moreover, we plan to apply our inference pipeline to electrophysiological measurement data in future research.

In the current study, we demonstrate that amortized \gls{sbi} can be successfully applied to a stochastic experiment with a high-dimensional parameter space on the neuromorphic \gls{bss2} system.
Moreover, our approach is agnostic to the specific hardware platform and can therefore be readily applied to other neuromorphic systems or to simulated data.
 
\section*{Acknowledgements}
\addcontentsline{toc}{section}{Acknowledgment}

We thank Jakob Huhle and Titus Zangemeister for their contributions to the SBI pipeline and experimental investigations, which paved the way for this study.

This work has received funding from
the EC Horizon 2020 Framework Programme
under grant agreement
945539 (HBP SGA3),
the EC Horizon Europe Framework Programme
under grant agreement
101147319 (EBRAINS 2.0),
and the \foreignlanguage{ngerman}{Deutsche Forschungsgemeinschaft} (DFG, German Research Foundation) under Germany's Excellence Strategy EX 2181/1-390900948 (the Heidelberg \mbox{STRUCTURES} Excellence Cluster).

We used several AI tools (PaperPal, ChatGPT and Gemini) to correct grammar and spelling as well as for rephrasing text.
All outputs were checked by the authors and they maintain full responsibility for the intellectual content of the manuscript.
 
\section*{Author Contributions}\label{sec:author_contributions}

We provide contributions in the \textit{CRediT} (Contributor Roles Taxonomy) format:
\textbf{JK}: Conceptualization, methodology, supervision, software, visualization, investigation;
\textbf{EM}: Conceptualization, methodology, supervision, software, resources;
\textbf{JS}: Conceptualization, methodology, supervision, funding acquisition;
\textbf{all}: writing --- original draft, writing --- reviewing \& editing.

\printbibliography

\end{document}